\documentclass[conference]{IEEEtran}
\IEEEoverridecommandlockouts
\usepackage{cite}
\usepackage{amsmath,amssymb,amsfonts}
\usepackage{algorithm,algpseudocode}
\usepackage{graphicx}
\usepackage{textcomp}
\usepackage{xcolor}
\def\BibTeX{{\rm B\kern-.05em{\sc i\kern-.025em b}\kern-.08em
    T\kern-.1667em\lower.7ex\hbox{E}\kern-.125emX}}
\begin{document}

\title{Computer Vision Based Parking Optimization System\\
}

\author{\IEEEauthorblockN{Siddharth Chandrasekaran, Jeffrey Matthew Reginald, Wei Wang, Ting Zhu}
\IEEEauthorblockA{Department of Computer Science and Electrical Engineering\\
University of Maryland, Baltimore County\\
Email: \{siddharth, jeffrer1, ax29092, zt\}@umbc.edu}}

\maketitle

\begin{abstract}
An improvement in technology is linearly related to time and time-relevant problems. It has been seen that as time progresses, the number of problems humans face also increases. However, technology to resolve these problems tends to improve as well  \cite{b1}. One of the earliest existing problems which started with the invention of vehicles was parking. The ease of resolving this problem using technology has evolved over the years but the problem of parking still remains unsolved. The main reason behind this is that parking doesn’t only involve one problem but it consists of a set of problems within itself. One of these problems is the occupancy detection of the parking slots in a distributed parking ecosystem. In a distributed system, users would find preferable parking spaces as opposed to random parking spaces. In this paper, we propose a web-based application as a solution for parking space detection in different parking spaces. The solution is based on Computer Vision (CV)  \cite{b2}\cite{b3} and is built using the Django framework written in Python 3.0. The solution works to resolve the occupancy detection problem along with providing the user the option to determine the block based on availability and his preference. The evaluation results for our proposed system are promising and efficient. The proposed system can also be integrated with different systems and be used for solving other relevant parking problems.
\end{abstract}


\section{Introduction}
In the current world which we live in, the number of people using vehicles is increasing day by day and with the invention of the automatic vehicles and the rising technologies revolving around it, it is safe to say that almost all individuals will own a manual or automatic vehicle in the near future. With the increase in innovation level comes the problem of updating the existing solutions to adapt to the new changes incoming or might be coming in the future  \cite{b4}. One such solution is the parking management system. In addition to that, time is of the essence for everyone. So, spending a substantial amount of time to figure out the parking slot and then heading towards the next task is unsatisfactory for almost anyone owning a vehicle. This issue comprehensively increases in frustration when there is a distributed parking management system. The number of parking blocks and slots one needs to check before parking substantially increases when it comes to distributed parking. 

A large majority of the existing solutions for occupancy detection is with the use of sensor-based devices  \cite{b5}. This kind of architecture is usually built into malls and large arenas with distributed parking slots. It usually comes with a one-to-one module i.e. each parking slot will be associated with one sensor. So, it becomes a linear relationship between parking slots and sensors. In addition to that, with each sensor comes the cost of setting it up, recharging it, maintaining the sensor in different locations or environments thus making it inefficient when it comes to large areas. This paper proposes an approach to detect parking spots  \cite{b6} with the help of fixed cameras in specific locations. It assumes that the spatial location of an n-number of cameras would be able to cover all the parking spaces and that there are preferable parking slots for each individual instead of the individuals being comfortable with parking anywhere within the distributed range of the parking system.\\

The proposed approach can be digested into the following steps:\\

Step 1: Detection of the parking spaces beforehand: The solution requires a manual step in which the technical support person for the solution needs to mark the parking spaces in each parking slot in the distributed parking system.

Step 2: Registration of the user to the solution: Since it is a web-based application, the user needs to register himself into the system to access the application. All the security aspects regarding registration will be taken care of in this module.

Step 3: Schedule Monitoring/Creation: Our approach is built based on a school-based prototype. Hence, the user can choose which classes he has enrolled in. 

Step 4: Dynamic video updates: The video captured by the fixed camera needs to be saved and updated into the folder from which the solution can pick it up and render it. 

Step 5: Slot detection: The user will be able to look at the current slots available and will be advised as to which slot he needs to park in as per his schedule [Step 3].\\ 

Even Though our prototype is based on a school or university based use-case, it is relevant to different real-world use-cases as well. For example, before an individual heads into a mall, it will be greatly beneficial to the individual to figure out which slots are available and based on the shop(s) he needs to visit, he can decide on the same. Our proposed solution is also open to integration with new components which can be useful in figuring out resolutions for different issues as well. The evaluation results which we received for our solution are promising and can only improve which is absolutely necessary. 

Another perspective of parking involves security of the vehicle, billing for the vehicle along with estimation of parking billing finances. All of these issues aren’t primary to the parking management system but they do exist and will continue to increase in volume and complexity as the system evolves. The unique nature of our application provides the ability to integrate sub-solutions which resolve the issues mentioned above into the solution and create an overall efficient parking management solution which deals with the primary as well as the lower-priority issues related to parking.

\section{Previous Work}

% \subsection{Maintaining the Integrity of the Specifications}

Parking detecting systems were built in the early 2000s using traditional hand-crafted elements. The majority of the current solutions rely on sensors to determine whether each parking space is occupied. Static traffic refers to car parking, which includes both short-term parking for passengers hopping on and off, as well as long-term parking in parking areas. Design and layout, insufficient regulation of land use indicators, fewer parking spaces in parking lots, fewer parking spaces in public buildings, and a severe road occupation phenomenon are all examples of static traffic difficulties. The above situation demonstrates that the parking problem is very acute. Design and layout issues, inadequate control of property use metrics, very few parking spaces in parking areas, very few parking areas in public buildings, as well as an extreme road occupation phenomenon are all examples of static traffic issues. The above situation demonstrates how serious the parking issue is.

As a result, enhancing parking spot detecting technology has risen to the top of the priority list for easing parking issues. Studies have discovered numerous parking space recognition methods to enable drivers to park and access the garage more swiftly with real-time parking space data in recent times, with the continuous development of parking space detection methods. Researchers have developed a range of parking space detection techniques, which can be split into fully automatic and semi-automatic detection techniques, as a result of the fast development in need for parking assistance systems in the last couple years. There is no need for manual intervention when using completely automatic parking space detecting systems. This device detects the required parking place and chooses it for you automatically. However, because all of the recognized parking space kinds are preset, there are severe limitations on the types of parking spots available inside. This system is unable to detect that it is outside of the specified range. The semi-automatic parking space detection approach, on the other hand, necessitates human-computer interaction during the parking space identification process to complete the recognition of available parking lots. As it has more information from the user, the semi-automatic approach may yield more reliable outcomes and utilize less computing resources than the fully automatic method. For instance, Toyota's IPA (Intelligent Parking Assist System) is a classic semi-automatic parking system that uses a human-machine interface demonstration to display possible parking bays on the rear-view camera image and allows the driver to alter the location of the parking spot using direction controls.

Despite the fact that semi-automated parking spot recognition has substantially reduced the difficulty of parking for drivers, many individuals still find it too difficult and confusing to use on a daily basis. As a result, academics have proposed a number of ways for totally autonomous parking space recognition. To execute a three-dimensional fitting of a car plane model, Kaempchen et al. \cite{b7} devised a stereo-vision-based method which uses a feature-based stereo technique, a template matching approach on a depth image, and an iterative nearest neighbor algorithm. Using Restricted coulomb energy (RCE) color separation, contour extraction using the least square approach, and inverse perspective transformation theory, Xu et al. \cite{b8} proposed a parking space labeling system. This approach, however, is solely based on parking space markings. It may be compromised due to poor viewing circumstances on the marks, such as spills, reflections, and partial occlusion from nearby vehicles. With the advancement of parking space sensing technologies, more researchers are exploring parking space recognition methods that rely on semi-automatic and fully automatic approaches, promoting the field's further advancement and growth. The methodologies and technologies for detecting parking spaces will be summarized and classified in this section. Simultaneously, it will study and describe the features of existing parking space diagnostic techniques, as well as the obstacles that parking space detection technology faces in this sector, in order to help future scholars better solve the parking space problem. All of the information can serve as a resource or aid in the detection of issues.

The resurrection of machine learning has switched the majority of computer vision-based technologies to the machine learning paradigm. There's not much research on deep learning-based smart parking systems in the research. This chapter includes much of the research on parking detecting systems that is available today. S. Nawaz et al presented a parking assistant system that relies on a smartphone. Their technology is based on a sophisticated sensor that detects a vacancy in the road parking space. They used Wireless internet beacons in the streets to detect vacant parking spaces. Their suggested system checks whether the motorist has started driving by measuring the rate of changes in beacons. They asserted that their technology outperforms GPC and Internet-based parking options. When compared to certain other location-based smart parking systems, this system is more energy efficient. This app runs on a smartphone and requires Wi-Fi to function.

Moreover, the actual vehicle's existence is not guaranteed. Likewise, Shau Ma et al used a combination of sensor outputs to locate parking and unparking operations. Their envisioned application makes use of data from various phone sensors. High-level human behaviors can be inferred using these probabilistic fusion algorithms. They say that by reducing the use of GPS, they can save electricity. They employed nine distinct signs to determine whether the vehicle was parked or not. Some indications are based on speed, while others are dependent on smartphone connectivity. They used a dataset they created themselves to train the algorithm. During parking and unparking activities, they obtained precision of 90 percent and 93 percent, respectively. The notion of employing gyroscopes in smartphones to detect unoccupied parking spaces was introduced by Villanueva F. Jesus et al.

Researchers present an empirical investigation to demonstrate the viability of employing smartphone sensors to locate parking spaces. They put their proposed method to the trial in a regulated college parking environment. To discover a parking place, Xavier Sevillano et al integrated data from sensors and a sensory stream. To see variation, they presented a pattern based recognition approach. They show a system for determining if a parking space is filled or vacant. For picture segmentation into empty and local, they used both local and global information. In order to obtain the visual description of the entire image, they used the Statistical features that included Gradient and the Gabor scatter plots. Researchers used Accelerated Robust Attributes to identify the traits of certain parts of the image (SURF). They used SVM and KNN classifiers to test their method.

Researchers trained the algorithm using on-site recordings and the UIUC Data Set. Unmanned aerial vehicles (UAVs) were used by Zhou et al to determine the amount of vehicles in the parking lot. Unlike many other detection difficulties, they were able to convert the car counting problem into pixel density estimates. For each vehicle, they used Scale Invariant Feature Transform (SIFT) feature extraction to extract spatial information. To create a codebook of characteristics, these extracted characteristics are clustered using K-Mean clustering. They used UAV photos acquired at Deakin University to train their algorithm. By exploiting the vehicle's camera on the smartphone, Giulio Grassi et al devised an algorithm which is based on the parked vehicle localization technique.

This technology is a client-server program in which a person's smartphone operates as the listener, receiving the visual stream and sending it to the web server, which maintains the actual data. When detecting vehicles, they used a Viola Jones-based object detection model. On the basis of positive and negative examples, the classifier learned the properties. This cascade classifier was trained using the San Francisco automobile dataset. Martin Ahrnbom et al proposed a new method for extracting features that is both quick and accurate. This method is very light and can be readily implemented on low-performance systems. They used integral channel characteristics with logistic regression for binary classification to classify the image into automobile parked and vacant. SVM was also used for supervised classification. The PKLot dataset was used to test their algorithm. Average based area under curve for regression models was found to be 0.98.

Harshitha Bura et al used a dispersed network of CCTV to provide autonomous parking support. Deep convolutional networks are used in their method to retrieve the high - level features of the parking space and vehicle.  They created a comprehensive framework for charging automobiles, locating vacant parking spaces, and determining the time a car was parked. Top view cameras, floor level cameras,  and edge devices make up their entire system, which provides a complete foundation for parking lots. They used the PKLot and CNRPark datasets to classify vehicles and attained a 99.5 percent accuracy for combined photos from these datasets. Deep convolution neural networks were used by Sepehr Valipour et al. To locate vacant parking places.

Researchers concentrated on the whole system's dependability and cost reduction. To make the total technology cost-effective, they used vision-based detecting techniques. Every parking camera may monitor multiple parking spaces, lowering the operational cost of each parking space. These cameras can withstand environmental factors, such as light intensity, weather and temperature. For slot image categorization, they used VGGNet-F, a simpler version of both the VGG network. They trained their system using the PKLot dataset.

R. Yusnita et al. suggested a system based on the image filtration processes, such as thresholding segments. Thresholding segmentation is an object categorization approach in which each pixel is replaced with white or black based on pixel data points. It also reduces background noise while collecting valuable information from a picture. However, because thresholding does not evaluate sensitive data, it merely removes the background and extracts the foreground based on pixel intensities, it also destroys valuable information. It determines which pixel is most closely associated with the intensities and changes the image accordingly. Making decisions based on thresholding is not a good idea because some obstructions can be mistaken for automobiles, resulting in occupied parking spaces, which can lead to erroneous recognition.

Ruimin Ke \cite{b9} et al. suggested a system for recognizing parking spots from a live video using an edge detection algorithm. This system can identify occupancy by parallelizing the process and taking into account the adaptability module for online data transfer, which is crucial for detection precision and accuracy. Their tracking system identifies automobiles and parking ground effectively in varied angle parking areas, however there is no idea of recognition and classification. Benjamin Kommey \cite{b10} et al. suggested a technique based on the morphological analysis and canny edge detection. Canny edge detection is a well-known method for extracting edges from images.

Whenever the image is a bit complicated, Canny fails in sophisticated edge extraction techniques because it produces broken lines. This method extracts the boundaries using canny and then uses morphological dilatation to fill in any holes or broken margins. However, in the event of canny, dilatation will not be successful in properly joining or filling the gaps. Another system based on a partial derivative edge detection technique was suggested by Chihping Hsu \cite{b11} et al. This system could distinguish between unoccupied and occupied parking spaces with 95\% efficiency, however there is no classification technique that distinguishes between automobiles and people. Pedestrians or other impediments may obstruct the parking space, causing the system to incorrectly recognize them as paved parking spaces. This system will be able to correctly classify all barriers and vehicles.

Paula Tătulea \cite{b12} et al. devised a foreground and background subtraction-based method. Background subtraction is a flawed method since it necessitates the usage of a foreground and background image that subtract from each other, resulting in changes based on extra objects laid over the background image. Any obstacle can be regarded as an object in this approach, but if it is inside the blob or coordinates, this same system will recognize that slot as filled, which is erroneous and will have a direct impact on system reliability. This approach is based on statistical changes, in which image pixel brightness is subtracted and a shift is produced. It is easily influenced by color brightness or light, both of which reduce accuracy. The strategies that have been established so far in the subject of parking spot detection do not give a perfect method for implementation in practice.

Background subtraction is a flawed method since it necessitates the usage of a foreground and background image that deduct from each other, resulting in changes based on additional objects laid over the background image. Any obstacle can be regarded as an object in this approach, but if it is within the blob or coordinates, the system will recognize that slot as filled, which is erroneous and will have a direct impact on system accuracy. Background subtraction (BS) is a frequently used approach for using stationary cameras to generate a foreground mask (i.e., a binary picture comprising the pixels relating to active objects in an image).

Even as name implies, BS computes the foreground mask by subtracting the current frame from a background model that contains the static component of the picture or, more broadly, everything that may be called background based on observed scene's features.

Regarding solving the issue of parking in congested regions, various approaches and techniques have been offered. Ming-Yee Chiu et al. devised a method for calculating the pool of available parking spots by measuring the cars at the checkpoint \cite{b13}. These induction loop sensors are installed beneath the road surface to do the counting. While sensors were less expensive, they have been less impacted by environmental conditions, and detected precisely, their installation was problematic and resulted in road damage. Also it was difficult to keep it running in the event of a breakdown \cite{b14}. Furthermore, the precise locations of free parking places cannot be identified because the counting system is only capable of recording the amount of cars going through checkpoints \cite{b15}.

Another detection methodologies are based on the employment of ultrasonic, infrared, and microwave sensors for object tracking \cite{b16}. Each and every parking space has these sensors installed beneath it. Wan-Joo Park et al. suggested that vehicles be equipped with ultrasonic sensors to seek for available parking spaces. These same sensors in this system are easily affected by weather conditions such as snow, temperature, rain, and a strong air breeze. Vamsee K. Boda et al. presented alternative method smart sensor nodes. This approach was less expensive, because it relied on wireless sensor nodes installed at key locations such as lane turns, parking lot entry and departure points. The difference between incoming and exiting cars can be used to compute the overall number of cars in the parking lot \cite{b17}.

The other types of detection systems are shown to be based on visuals. The entire parking lot available for parking may be evaluated through the cameras using vision-based technologies, the data is then analyzed, and the result provided will indicate the actual figure and position of parking slots. According to Zhang Bin et al., vision-based parking spot detection systems are simple to install, relatively cheap, and the detector may be quickly changed to meet specific needs. Furthermore, the information gleaned from photographs is quite valuable. Unfortunately, one of the vision method's flaws is that the precision is heavily dependent on the camera's placement. Regarding the parking spot monitoring system, Thomas Fabian suggested an unsupervised vision-based method. This suggested approach requires fewer video frames per minute and also has a low computational complexity.

The main issue in picture recognition, they argue, is shadows and occlusions \cite{b18}. Sophisticated clustering techniques are utilized for unsupervised learning. The vision-based parking spot recognition system, according to H. Ichihashi et al., has been mostly affected by climate and illumination conditions, such as drops of rain landing on the lens of the camera during intense rainfall. Lighting conditions that are both high and low. As a result, the cameras are typically employed to detect automobiles in indoor parking spots rather than outside parking spaces \cite{b19}. R. Yusnita et al. described a technique in which each parking lot was individually marked with a brown hue circular patch. Whenever the system is first started, it searches for a circular form in each area; if a patch is found, this area is considered vacant and the vehicle is displayed \cite{b20}.

Whenever cars block the patches, the system thinks that the slots are occupied by cars. This technique worked effectively for managing the parking area, but it failed miserably in severe snow and rain. N. True suggested a method for detecting parking spaces that combined color histogram and vehicle attributes detection \cite{b21}. Mostly in an outdoor parking area, Najmi Hafizi presented an image-based approach for detecting open slots. For gathering photographs of the parking lot, a low-resolution web camera is used, which considerably minimizes the cost.

These photos are standardized, and then a couple of ROIs are added to each partition of the parking space, increasing the accuracy of vehicle detection \cite{b22}. \cite{b23} describes an image processing approach for detecting whether a parking division is available or allocated by capturing the dark circle printed on the parking space and processing it. In \cite{b24}, a car image is recorded as a reference image, and the other pictures are matched with the source images using an edge detection approach, displaying information about available and occupied spots. Several approaches for extracting features from photos have been presented. They have formulated and implemented a framework for an autonomous parking system in their study. These experimental data show that the proposed system offers remarkable accuracy when compared to state-of-the-art techniques.

We move on to discuss the proposed parking lot architecture in detail in Section 3. We then go on to brief about the proposed algorithm and proposed system in Section 4 and 5 respectively. Section 6 deals with the potential clients for our application and Section 7 elaborates on the services offered by our application. In Section 9 we go onto discuss the experimental results followed by the future work in section 11. Finally we conclude in Section 12.

\begin{figure}[h]
\centerline{\includegraphics[width=0.45\textwidth]{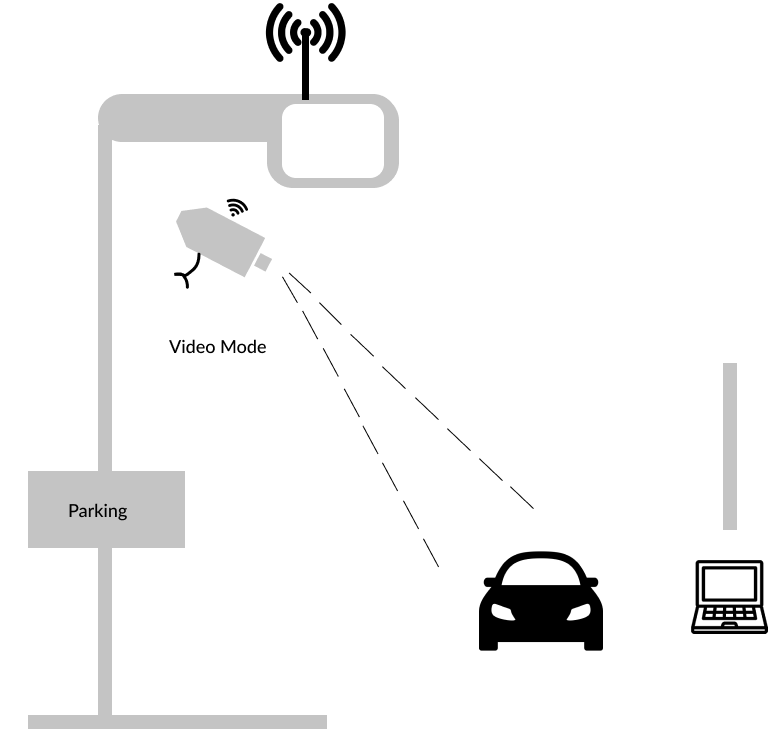}}
\caption{Architecture}
\label{fig1}
\end{figure}

\section{Architecture}

Our proposed system works based on the video captured by the parking lot cameras. The cameras in the parking lot get a wide angle shot of the complete parking space we require to process. The live video is then transferred to a database in intervals of 10 seconds using the internet service. This video is then processed by our algorithm to detect the number of available parking slots with respect to the total number of slots. Initially the slots are marked manually which is a one time manual activity but the work is negligible considering the benefits it provides. We have used the Django framework to develop our web application using python.

\subsection{Django}

Django is a well-known Python web application framework. It operates on the "batteries-included" principle. Batteries-included is based on the idea that typical web app functionality must be included in the framework rather than as separate packages. The Django framework, for instance, comes with URL routing, authentication, a template engine, database schema migrations and an object-relational mapper (ORM). In contrast to the Flask framework, which uses a separate module called Flask-Login to provide user authentication. The two approaches to framework development are scalability and batteries-included approaches which are two different approaches widely used. Innately, neither ideology is superior to another.

Django is a web framework built by skilled developers that takes pretty good care of the heavy lifting so you can focus on developing your app instead of reinventing the wheel. It is indeed open source and free, with a vibrant and active group, a variety of paid and free support options and excellent documentation.

Django also takes care of our application as it helps to write complete, secure, versatile, scalable, maintainable and portable applications.
A web app listens for requests from the browser window in a standard data-driven webpage or any other client. When an app receives a request, it determines what is required based on the URL and perhaps data available from the GET or POST payload. This may next access the data to write or read the information from a database or do other operations to complete the query, based on what is necessary. The program will then send a reply to the web browser, commonly by putting the acquired data into templates in an HTML template and subsequently constructing an HTML web page for the browser to render. That code for each of these processes is usually separated into separate files in Django web - based applications.

To access this web based application clients first need a username and password which can be generated using the registration screen. After registering the user needs to login to the system. Our application is tailored to the needs of students as we forecast that this application would become a hit in the college community as students find it difficult to find parking slots on campus. Thus we have tailored our application to get the student class schedule so that the students can find parking spaces based on their class schedule and location. After giving their necessary class schedule students can click on find parking when they need to find a parking slot. On clicking the find parking button our application runs the algorithm to find the number of available slots in the parking slots near to the students class and give them the required count. This is just an overview of our proposed approach. In the further Algorithm section we discuss in detail about how we detect and identify the number of available spaces in detail.

\section{Algorithm}

The algorithm involved behind our proposed approach can be split into different sets of algorithms working together which are described as subtopics:

\subsection{Algorithm 1: Video Capture Algorithm}

\begin{algorithm}
\caption{Video Capture}
\begin{algorithmic}
    \For{video in videos:}    
        \For{frame in video:}  
        \State $Frame\_values = detect_motion(frame)$
        \State $updateDB(Frame\_values)$
        \State $updateUI(Frame\_values)$
        \EndFor
    \EndFor
\end{algorithmic}
\end{algorithm}

Since a video is a set of frames and the approach is built on a video sequence which means the video is of fixed-length, the solution captures each frame one by one and determines the occupancy slots. This algorithm is programmed with the help of open source computer vision packages built for python3.0.

\subsection{Algorithm 2: Space identification algorithm}

\begin{algorithm}
\caption{Space identification}
\begin{algorithmic}
\State $action\_quit \gets False$
\State $Marked\_lines \gets []$
\State $Cnt\_of\_slots \gets 0$
\While{not $action_quit \neq False$}
\State $Marked\_Lines \gets Marked\_Lines+mark\_contour(cursor)$
\If{$formrect(Marked\_lines) \neq True$}
    \State $Cnt\_of\_slots \gets Cnt\_of\_slots+1$
    \State $storepixels(Marked_lines)$
    \State $Marked\_lines \gets []$
\EndIf
\EndWhile
\end{algorithmic}
\end{algorithm}

The technical support person needs to mark the slots present in the parking space with the help of an image from the parking space. With the help of cursors, the person would be able to manually mark the individual parking slots one-by-one.

\subsection{Algorithm 3: Occupancy detection algorithm}

\begin{algorithm}
\caption{Occupancy detection algorithm}
\begin{algorithmic}
\State $action\_quit \gets False$
\State $Marked\_lines \gets []$
\State $Cnt\_of\_slots \gets 0$
\While{not $action_quit \neq False$}
\State $Marked\_Lines \gets Marked\_Lines+mark\_contour(cursor)$
\If{$formrect(Marked\_lines) \neq True$}
    \State $Cnt\_of\_slots \gets Cnt\_of\_slots+1$
    \State $storepixels(Marked_lines)$
    \State $Marked\_lines \gets []$
\EndIf
\EndWhile
\end{algorithmic}
\end{algorithm}

The solution needs to take care of two cases when it comes to occupancy. When the slot becomes available, it should be indicated with green and when the slot becomes busy, it should be indicated with red. The intensity of the pixels at each frame in the video help to determine the occupancy. This algorithm is programmed with the help of open source computer vision packages built for python3.0.

\section{Proposed System}

Our proposed system is a web-based application which provides the user the option to look into the occupancies for each parking block without the help of an external technical user. It eases the comfortability for the individual using the application. The proposed system also focuses on a dynamic rather than a general model for the taken problem because the same solution works differently for different users based on their preference(class-schedule). Therefore, it handles the multiple user scenario as well. One of the major questions which arise when a common solution is devised is the security as well as the parallel use-case for the solution. The proposed solution has security intact because the authentication factor uses the in-built Django module for authentication where the passwords are stored in hashed formats rather than plain-text[25]. 
The parallel use-case scenario is dealt with on the basis of having class schedule-based executions of the solution i.e. the solution works differently for different class schedules which infers that the solution works differently for different individuals. 

The resolution of the problem is resolved primarily with the help of computer vision libraries. The function of enabling computers and systems to derive meaningful information from digital photos, videos, and other visual inputs is referred to as computer vision.The integration of computer vision libraries into the solution provides the capability to mark the parking slots, store the intensity of the pixels in each slot, render the video frame by frame and detect occupancies based on difference in intensity. The open-source Computer Vision libraries enable the solution for occupancy detection to work well. The parking spaces are detected with the help of numpy arrays which come along with the Computer vision packages when you install it. The first step to identify the existing parking slots requires manual intervention which is not costly since it is a one-time effort for each parking block and needn’t be repeated unless the parking slot is renovated or redesigned which rarely happens. Once the parking slots are identified, the locations of the parking slots are stored in a file in JSON format so that they can be passed over to another computer when the solution is transferred over as a whole. This further signifies the individual nature of the project. This step can be considered as a prerequisite step before using the solution. 

The second part of the solution involves registration of the user into the system. Since the solution was developed with multiple users in mind, the registration component is integrated into the solution. An user can register himself with the help of an username and password and login into the solution to look at his schedule. While registration, the security aspect is taken care of by hiding the password as the user types it in and when the user tries to login to the system as well, the password will be hidden from the eyes of the user. For password storing, the default Django authentication module has been built to the solution so that hashing and other security-driven aspects of the authentication module are dealt with. Since the strength of a password is also important when it comes to maintaining the security[26], users will only be allowed to set strong passwords which we will discuss in the implementation section of the paper. 

The third part of the solution revolves around the user editing or choosing his schedule. The user has the ability to edit and choose his class schedule so that the solution can provide suggestions on which block he needs to visit so that he doesn’t waste time by walking or choosing for random slots in different blocks which are far away from the class. While the user logs into the system for the first time, he will not be shown any classes as per his schedule which is expected. When he edits his schedule, he will be able to see all the classes which are present in the current semester and choose which and all he is currently enrolled into with the help of a Yes/No question. Once he saves his selections, he will be able to see his schedule on his home page. He can change his schedule at any point of time because he might modify his classes or add some classes into his schedule as per his interest and needs which quite often happens. 

The fourth part of the proposed system is the vital one. The solution when given a video, it will detect the number of occupied and vacant slots and provide the result to the user with respect to the time when the user tries to check it. Once a video is recorded, it needs to be placed into a location mentioned by the solution so that it will be picked by the solution and rendered and show suggestions to the user. There might be confusion regarding the question of what if multiple videos are recorded. This issue has been dealt with by fetching the latest video from that particular location however the system which places the video into the folder needs to input that it has placed the video into the database table which stores the video list. A database table is being used over here so that the history of the videos are maintained and if the video location is hacked or incorrectly deleted for any reason, we will definitely have a back-up. Thus, the database table provides a safety option. After the video is fetched by the solution, the solution will have two aspects of the prominent solution in its hand. One will be the location of the parking slots in that particular location and another will be the video which it needs to render. As the solution renders the video frame by frame, with the use of the first aspect (location of the parking slots as JSON file) , it will be able to determine whether the slot is occupied or vacant and update the user screen as well as the backend. Since it renders the video frame by frame , it can also be used for detecting trends or patterns of the car parking slots which will be as part of future work. Once the video is rendered completely, the final status of the blocks of the parking slots will be mentioned to the user and the solution will provide suggestions as to which slot the user needs to take based on his class schedule. Currently, the mapping between slots and blocks is built using a database model but it can be converted into a geographic location model using google maps API which we will discuss further in the future [27]. 

The final part of the solution is the user experience. Once the solution finds the slots, it will list down all slots which are open in each block and the user will be able to decide which slot he needs to go. Also, since the solution provides a suggestion as to which slot will be helpful for the user, the user can only read the final suggestion and start his way to the parking slots. This will be helpful for students who are in a rush to the class and they wouldn’t want to miss a few precious minutes of a lecture.

\section{Clients}
Finding a parking place has become a daily concern in all of our lives in today's modern world. Everyone, I'm sure, has dealt with it at least once a week. The rising number of automobiles and, in particular, inadequate management of parking allocation systems are the causes of these parking challenges.

Urban areas are always changing. Citizens and tourists may struggle to find parking places if current and accurate data is not available, resulting in waste of time, congestion, emissions, and annoyance. As per ITS America's Market Analysis, vehicles searching for parking spots are responsible for roughly 30 percent of traffic congestions. Around one million gallons of oil are wasted every day around the world as a result to find a space for parking. As a result of the ever-increasing traffic jams and the unpredictability around available parking spaces and payment, a Smart Parking system is now required. A parking guidance technology that optimizes public parking space usage, increases parking operational efficiencies, and improves traffic flow. Smart Parking solutions are intended to provide motorists with an end-to-end resolution for their trip, eliminating the need to look for parking, calculate costs, or calculate travel time. These solutions when implemented would optimize the parking spot usage which in turn reduces the traffic which is created in finding a parking spot. These solutions in turn reduce pollution and enhance the user experience in finding a parking space. As we all have taken a cab to certain places just because it would be hard to find a parking spot for our vehicle. These problems would be solved by using our application. It also decreases the stress involved in finding parking spots. Hence people with all these problems in addition to that of students correspond to a huge set who are the potential clients for our web application.

\section{Services}

In this section, we will discuss the server components involved in the solution along with the user interface perspective of the solution as well. All the different components of the server along with its implementation will be discussed in this section. 

\begin{enumerate}
  \item \textbf{Registration of the user:} Any new user needs to register to login into the system and use the solution.
  \item \textbf{Class schedule registration:} The user needs to edit his class schedule based on the classes available. 
  \item \textbf{Parking slot availability detection:} The detection of the vacant as well as the occupied slots as well as a counter which maintains the count of each. 
  \item \textbf{Class schedule based detection:} The solution proposes suggestions based on the schedule which the user is currently enrolled in. 
\end{enumerate}

\begin{figure}[h]
\centerline{\includegraphics[width=0.45\textwidth]{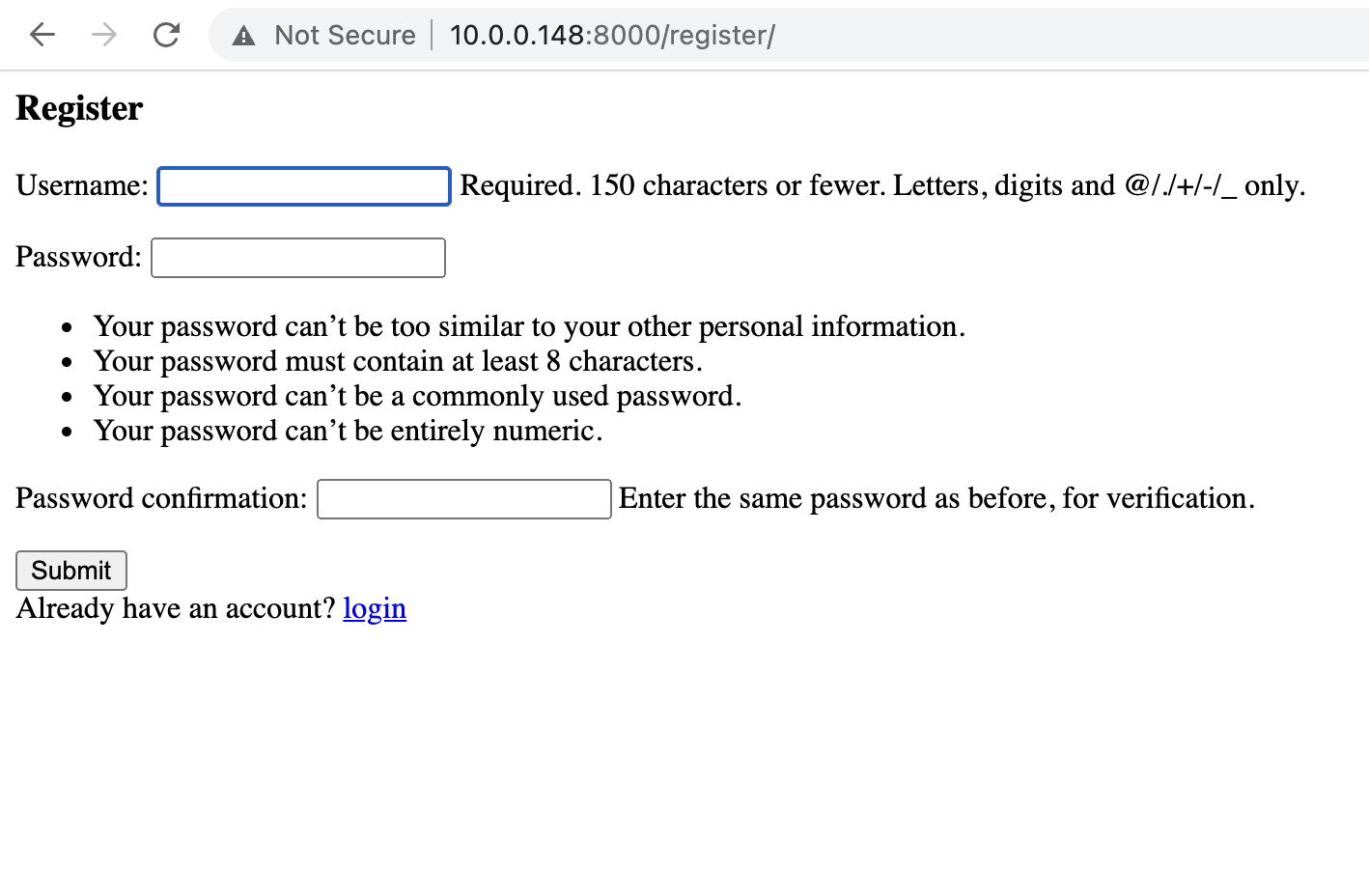}}
\caption{Registration Screen}
\label{fig2}
\end{figure}

\subsection{Registration of the user}

Any user can register into the solution with the help of an username and a password. The strength of the password[28] is taken care of with the help of the following metrics:
\begin{itemize}
    \item The password shouldn’t be similar to the username.
    \item The password must contain at least 8 characters. 
    \item The password cannot be a common password.
    \item The password cannot contain all numbers. 
\end{itemize}

Along with the security of the password, since the authentication module uses the inbuilt Django 
User accounts, groups, permissions, and cookie-based sessions are all handled by the authentication system.The final promising part of the Django authentication module is promising since it allows cookies to be enabled which means that the solution will be able to keep track of the browser which the user uses to login and if he tries to use the same browser without clearing the cache, he will be able to login without the help of his password[29]. This increases the easability section for any user. Since the solution uses the inbuilt Django module for authentication, it is clear that the hash of the password is stored into the django inbuilt tables which are written for storing the hashed password and it is assumed that the hash-collision rate of the hashing algorithm which has been devised by the inbuilt algorithm is pretty low so as to protect against brute-force attacks. The registration module can be further enhanced with the help of having a default list of users and passwords which can be traded into the solution with the help of database scripts which is open to discussion. 

An user who registers into the solution shouldn’t forget his password since he will not be able to retrieve it back. By default, Django doesn’t provide a password-recovery mechanism however the solution can be modified to build this as part of the next installment as well. The number of users who will be able to login into the system is high and is in accordance with the size of the database tables which store the hashed passwords along with the username and other relevant information as well. Since it uses the default module, it is assumed to be pretty high. Once the user registers into the solution, if he inputs the wrong username or password, he will be provided with the same error message and he can try again until he is able to get it right. 

\begin{figure}[h]
\centerline{\includegraphics[width=0.45\textwidth]{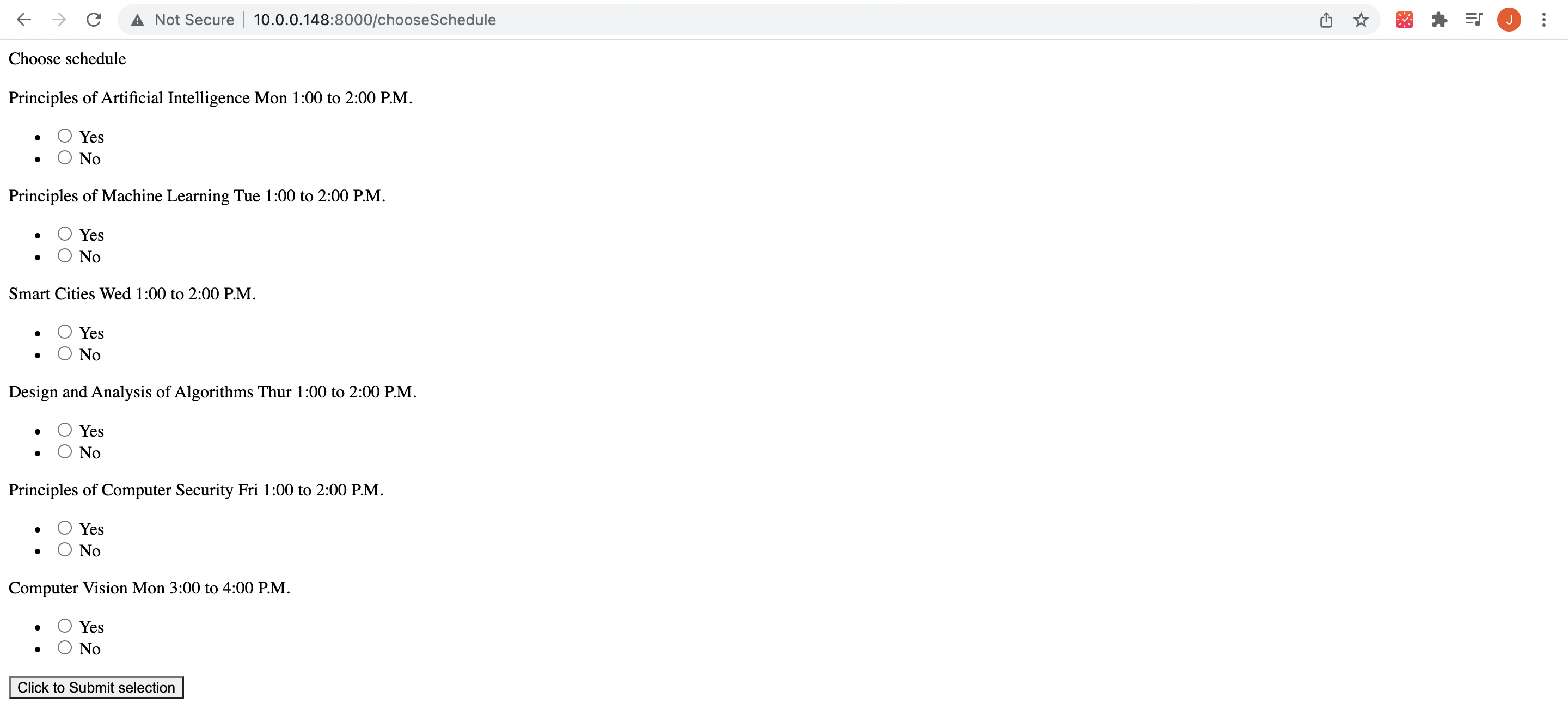}}
\caption{Class Schedule Registration Screen}
\label{fig3}
\end{figure}

\subsection{Class schedule registration}

Once a user registers into the solution, he can login back into the solution to check the number of slots in each block or edit his schedule. Since we have implemented cookie-based login as well, the user needn’t login again if he uses the browser and doesn’t clear the session or if he doesn’t use incognito mode. After the user successfully logs into the solution, he will be able to edit or choose his schedule based on the classes provided. The user will be shown a menu [Fig.3] which shows which classes are available and he will be able to choose which classes he is currently enrolled into with the help of Yes/No questions. He can edit his schedule at any point of time. This part was added because it is common in a university setup that the class schedule for a specific user isn’t certain for an entire semester and is susceptible to change. The user can click on “Edit my schedule” to edit his schedule and he will be able to see his schedule in his home-page. 

The timings of each class is also mentioned in the schedule page and is trivial to the solution because it provides suggestions based on the current time and the time relevant to the classes which the user is enrolled in. If two classes have similar timings, the same suggestion will be provided to both the classes. If two classes have different schedules, then based on the current time, the suggestion will be provided by the solution to the user. The time-zone of the schedule is also important to the solution and is assumed to be in Eastern Standard Time(EST).

\begin{figure}[h]
\centerline{\includegraphics[width=0.45\textwidth]{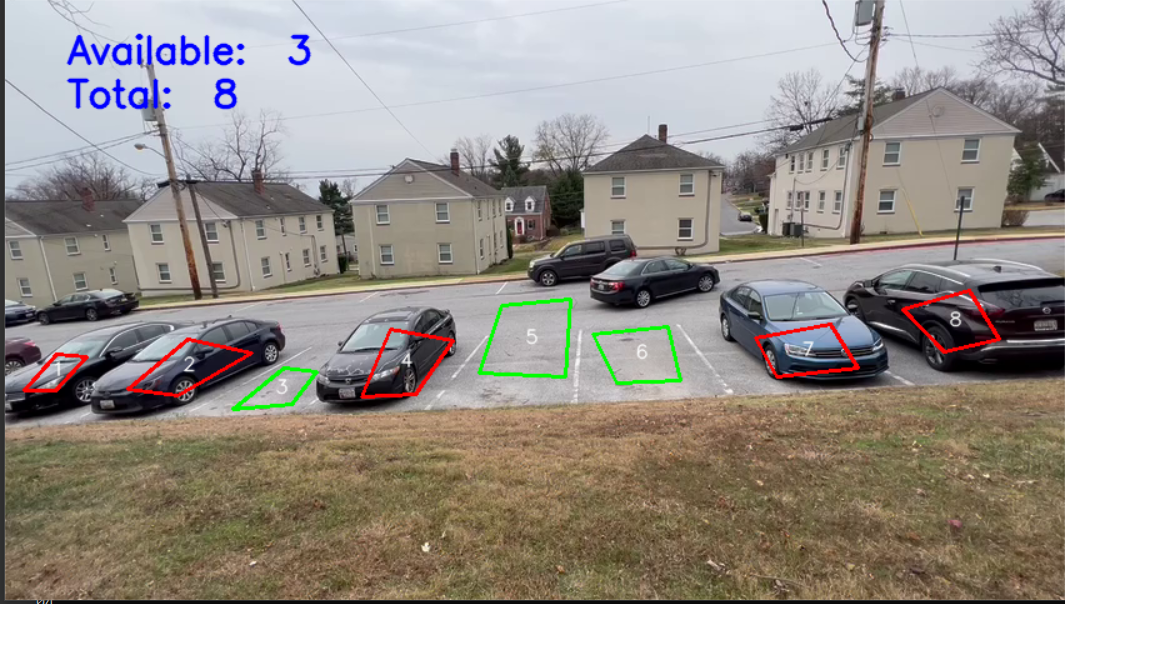}}
\caption{Finding Available Parking Slots}
\label{fig4}
\end{figure}

\begin{figure*}[h]
\centerline{\includegraphics[width=\textwidth, height= 8.5cm]{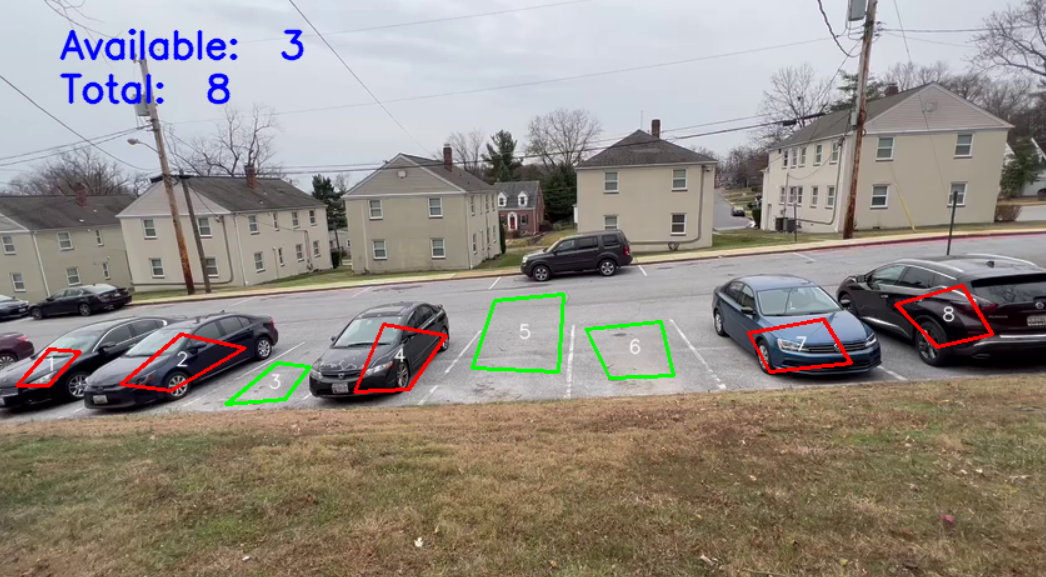}}
\caption{Snapshot before parking at slot 8}
\label{fig6}
\end{figure*}

\subsection{Parking slot availability detection}

Since we have taken a distributed parking management system, the parking slot availability detection is performed on each block within a distributed parking management system. For each block within the system, video is recorded and stored into the shared video location and the latest is picked up by the solution. The user or system which places the video into the shared location is assumed to input the video-name into the database table for the videos so that the solution will be read the names and pick it up from the location. This approach was taken so that the history of the videos can be stored in the table along with the location for safety purposes[30]. The link between the parking slots JSON file along with the video-names is also registered in the table which is also one of the reasons to follow this approach. It is ensured that an incorrect video-image matching isn’t performed. 

For each slot , for each video, the video is rendered frame by frame and based on the input JSON file for that corresponding video file, the slots are detected and will be validated if there is any modification in the intensity of the pixels as to how it was earlier. If there is a modification in the intensity and there are more pixels with the gray color are noted, it means that the particular slot is vacant and if not, it means that the particular slot is occupied. With this information, the solution is able to detect the availability factor for each parking slot in each block. Once a video is rendered over, the next video is taken and so on and on. The solution stops working only when all the videos are rendered over. The user will be able to see real-time as to how the occupancy slots are changing in color from RED to GREEN as they change from occupied to available and they will also be able to see a counter which counts all the slots available vs all the slots in total. 

As to the technical part of this particular module, numpy arrays are used to store the contours of the parking slots and they store the pixel density of those slots as well. (pixel density is more related to the RGB level of each pixel within that particular slot). The numpy arrays are iterated one by one based on the JSON file and the check on occupancy is performed on each slot and is mentioned in the screen as well.If there is a change in the occupancy property for each slot, the numpy arrays are updated accordingly. The video rendering is performed using a while loop until there is no frame to traverse over. Since every video is traversed frame by frame i.e. image by image, it ensures that nothing is missed at any point of time. Ensuring that no frame is left over with respect to time will be helpful in performing analysis on the parking data and try to figure out trends or project new slot occupancy rates as well.

\begin{figure}[h]
\centerline{\includegraphics[width=0.45\textwidth]{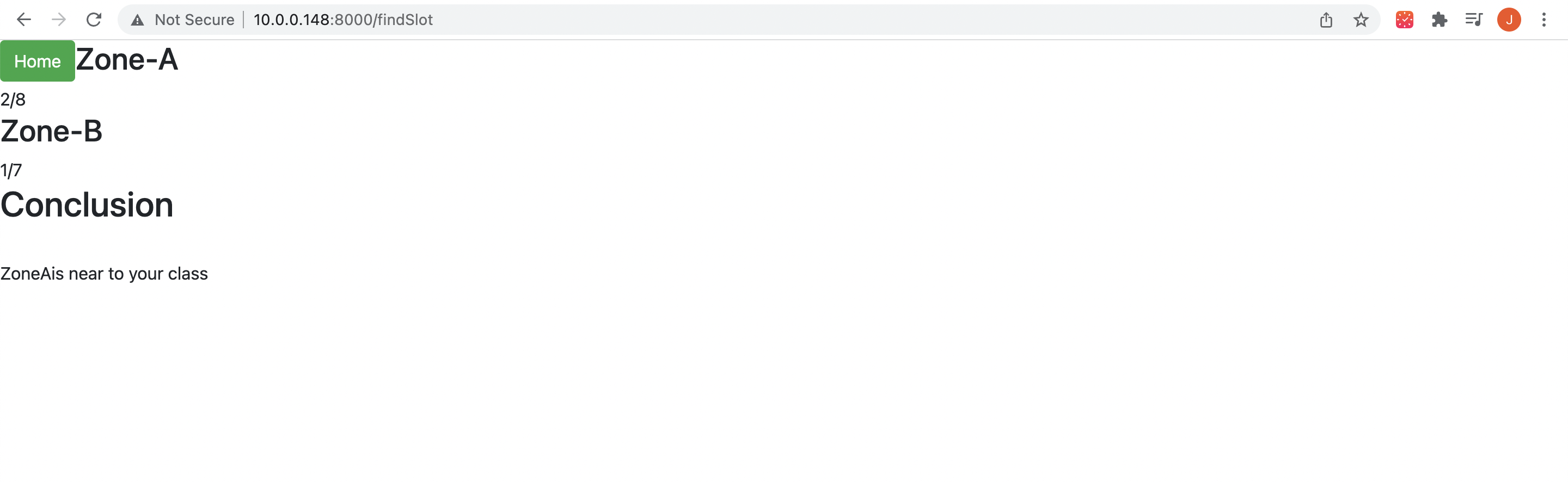}}
\caption{Number of available spots based on class schedule}
\label{fig5}
\end{figure}

\subsection{Class schedule based detection}

It is known that the user will choose his schedule when he logs into the solution. This is done so that the solution will be able to provide suggestions to the user as to which slot will be beneficial for him to traverse into when he makes his way into the parking garage.The schedule of the classes will be different since the same user will have the option to be enrolled into multiple courses at the same time within a semester. The timings of each class is assumed to be fixed throughout the semester and since most of the universities work with a fixed schedule throughout the semester unless any havoc happens, the assumption can be considered as a good one. 

As and when the user clicks the “Find slot” option, he will be able to see the recordings of the videos of each block and see how the number of open slots change in each block as well. He will be provided with the option to see all the blocks involved within that particular distributed parking management system. This will be helpful if he has a preferable slot for some reason in each block and he can check if that particular slot is available or not. After the user clicks on “Find slot”, the implementation occurs as discussed in the previous subsection. Once all the videos are rendered over, the solution does a match between the current time when the user uses the solution and the schedule of the user and makes suggestions on top of it. Few of the suggestions are taken for this particular prototype and an n-number of suggestions can be built into the solution for various cases but a generic type of algorithm to devise these suggestions is a limit of our proposed solution.

\begin{figure*}[h]
\centerline{\includegraphics[width=\textwidth, height= 8.5cm]{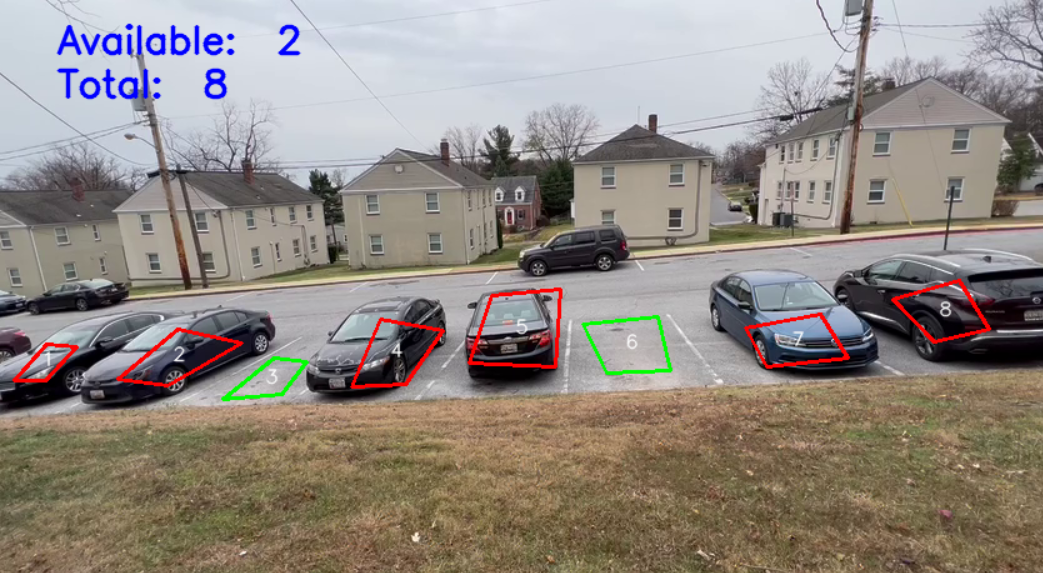}}
\caption{Snapshot after parking at slot 8}
\label{fig7}
\end{figure*}

\section{Test Bed}

We tested our application using two different parking lots. Let's name it as parking lot 1 and parking lot 2 for convenience. We mounted a camera on a higher level to get a wide angle view of the parking spots so that our solution is able to precisely detect the number of available parking spaces. The parking lot 1 consisted of 8 parking spots covered by the camera. We drove a car across the parking space to park in different spots which were accurately identified by our system that reflected the exact count using the front end. The parking slot 1 had 5 vehicles parked in a stationary position which was also detected successfully by our system. The parking lot 2 had a total of 7 parking spots covered by the camera. As seen in the parking lot 1 we drove a car similar to that of lot 1 in lot 2. Our application successfully detected and reflected the count of parking spots using the front end. These lots were extensively to test our application and the evaluation methods along with the results and discussions if discussed in brief in the following sections.

\section{Evaluation}

For this particular prototype, we have evaluated our solution with two parking blocks which make up the distributed parking management system. Each parking block has a different number of slots and this has been taken into consideration for similarity issues. The evaluation results for both the slots have been recorded and mentioned below. The classes have also been split such that 3 of them are close to slot-A and 3 of the remaining are close to Slot-B. 

For the first slot, the number of slots taken are 8 and when the video is rendered over frame by frame, the solution is able to detect the change in occupancy and change the color of the marked slots as well. In addition to that, it is able to make changes to the counter which records the number of available slots along with the total slots. It needs to be ensured that the number of total slots is assumed to be the total number of slots which are captured as part of the manual marking process which stores the location of the slots in JSON format. 

For the second slot, the number of slots taken are 7 and similar to the first slot, as the video is rendered over frame by frame, the solution is able to detect the change in occupancy and modify the color of the input marked slots. With an enhancement in it is that the solution is also able to make modifications to the counter which is used to record the number of non-occupied slots with respect to the number of total slots available. The counter is also shown on top of the screen during video rendering which helps to see the counter modified in real-time. 

As part of the final component of the solution, we evaluate the solution with different time-cases and see how the solution reacts to it. For a majority of the cases, the solution is able to determine which slot is near for the particular user based on his schedule and provide a suggestion. The suggestion also depends on the current time when the user asks to provide a suggestion as well. The solution assumes a few suggestions to be true and there is a limit on it as well. So, as part of further implementations , a generalization of the suggestions can be embedded and decisions can be made accordingly.

\section{Discussion}

As per the evaluations, it is seen that the proposed solution is able to detect the number of occupied and unoccupied slots and modify the counter as well. It is also able to provide suggestions to users based on the user schedule and the current time after the user clicks the “Find slot” option. As per the claims proposed by the paper, it is seen that the computer vision libraries integrated with the Django framework is able to provide an overall solution which detects the occupancies of parking slots in a distributed parking management system. 
The primary reason behind the proposed approach was to resolve the parking occupancy problem which is prevalent in the modern era and it is able to provide a solution for it according to the existing prototype. 

The solution is limited to work with fixed videos and is not able to run a real-time video rendering i.e. it cannot perform occupancy detection on the fly. However, the number of seconds which are missed depends on the time taken to place the video into the location and is assumed to be low and is negligible. Also, since the parking slot movement for a distributed parking management system will not be highly dynamic (since we have seen in his malls that the number of parking slots which change periodically is pretty low), the delay in the video being placed in the parking slot can be ignored. 

The solution is also limited to the factor that the camera which records the slots need to be fixed in a manner such that all the slots in that particular slot are covered. This limitation is overcome by assuming that this particular limitation exists for any solution which is derived for parking applications and so it can be considered as a requirement for any parking applications or parking solutions. The fixed cameras would need to capture the video in a considerably moderate video quality which is more than sufficient in normal conditions. 

Another limitation is that the proposed approach was not tested with a night-time approach but since it works the same for different parts of the day, it is assumed that it will work in a similar manner during night time. However, the cars would still need to be visible during night time i.e. extremely dark and black cars might cause an issue. Hence, taking all of these into consideration, the proposed approach is able to solve the problem of parking occupancy detection and provide up-to-date information on the same and since the module exists as an independent component, it can be integrated with other ad hoc solutions for any other parking use-cases as well.

\section{Future Work}

The problem of computer vision based system still remains unsolved and security problems \cite{b31}. In contrast to the existing work which can only detect the number of available parking spots on roads, the solution can be extended to detect parking lots occupancy in various regions such as grass, brick and mud parking spots. Unlike other solutions where only the rectangular parking spots are detected our solutions do not yield with that problem. Machine learning techniques along with the neural network approach can be used to automatically detect the parking lots at the first which we leave it as a future work. In addition, it is also possible to combine the proposed approach with the existing wireless and sensing  techniques \cite{b32, b33, b34, b35, b36, b37, b38, b39, b40, b41, b42, b43, b44, b45, b46, b47, b48, b49, b50, b51, b52, b53, b54, b55, b56}.

This application can now be accessed by any smartphone, computer or laptop that can connect to the internet. This web app can be extended to be available as an android and ios application which would increase the number of users drastically as it is always preferred to use an application in comparison to the web browser by the current generation. This solution can also be extended to give notifications for smart watches as they near the parking space. 

The web based application can focus to implement the allocation of specific parking spots to customers who have already registered using the online application. This gives a picture of the parking space in a much advanced fashion. In the future the application should be extended to concentrate on the security aspect of the parking space in addition to maintaining the intelligence of the software. Such as raising an alarm by detecting any abnormalities and thereby guaranteeing the security of the customers. Example if a person is standing near a car for a very long time with abnormal behavior or randomly peeping into cars can be detected and an alarm can be raised to the car owners and the concerned department to look into it which makes the parking spot a safer place.

The system can be further extended to automatically detect the number plates and thereby bill based on the amount of time a specific spot was used. This can be very useful as it can cut the manual labor to a great extent. This can be further extended to analyze the trends in the parking spot to design a better approach in much earlier fashion. Such as the system can be designed to show the trends of parking over the week and thus thereby learning how the parking slots are going to be occupied in the future weeks. Example in some places the parking slots may be occupied completely on mondays and wednesdays whereas it would be of moderate use on the other days of the week. Using these trends we can anticipate these much earlier and take necessary actions to overcome these obstacles. These trend detection can also help to dynamically allocate the parking zone which can also be incorporated to the system.

Take an example of a situation where there are 2 slots allocated for students and 2 for the faculty members and we find a trend where the slots for students are full during the 1st half of the day and reduce as the day progresses while that of staff are slowly increase from the start of the day and start to become more than half full during the later half of day. Conventionally an additional parking lot must be created to solve this problem which is of very high cost. This application can be extended to solve this particular problem by dynamically allocation zones based on the trends so that the same slot can be shared with the students during the 1st of the day. This dynamic allocation is of very low cost compared to that of the conventional approach and is the best cost effective solution. 

This solution fails in places with very low visibility such as poor weather conditions snow, rain etc. The future work can be used to resolve these above mentioned drawbacks. We leave it to the future to test the application on indoor parking lots as the problem with the indoor parking spaces is, in these places the video is hindered by beams, pillars and low contrast markings. For the system to be better recognised accuracy is very important while detecting the available parking spaces.

Pricing engines can also be incorporated into this system that can price the parking spot based on demand and availability ratio and the trends can also be used to price the parking spots which would yield a much higher revenue to the organization. This system can also be used to detect violations on parking lots and can be used to automatically generate and mail the bills to the corresponding driver which would not require any manual labor to go out and look for violations in the parking space. 

Once these trends are implemented the application can further be extended to suggest optimal time to get a preferred parking spot for a user. The system can also be tweaked to send out notifications such as an alert to the user if there is a prediction that his preferred slot would be filled during his preferred time.

Finally the application can help users to detect their vehicle by mapping them using a map and showcasing that using an interactive graphical user interface so that it becomes easy for the users to find their respective vehicles.

\section{Conclusion}

Parking issues exist in the modern era and with the ever-increasing number of users who use vehicles, it is certain that these issues will continue to exponentially increase in the future. The proposed solution is able to resolve one of the problems - parking occupancy issue and is able to successfully do so for the distributed parking system as well. With the help of this along with few enhancements, the solution can be devised into various applications and save time for an individual who wants to park and improve the efficiency of a parking management system as well.

\end{document}